%% file: main.tex
\documentclass[letterpaper, 10 pt, conference]{ieeeconf}  

\IEEEoverridecommandlockouts                              

\usepackage{cite}
\usepackage{amsmath,amssymb,amsfonts}
\usepackage{algorithmic}
\usepackage{graphicx}
\usepackage{textcomp}
\usepackage{xcolor}

\usepackage{lipsum}

\usepackage{physics}
\usepackage{mathtools}
\usepackage{bm}
\usepackage{xfrac}
\usepackage{nicefrac}
\usepackage{tikz}
\usetikzlibrary{backgrounds, chains}
\usetikzlibrary{qrr.shapes.openrectangle}
\usepackage{optidef}
\usepackage{siunitx} 
\usepackage{multirow}
\usepackage{multicol}
\usepackage[caption=false]{subfig}
\usepackage{cancel}
\usepackage{url}
\let\labelindent\relax
\usepackage{enumitem}

\makeatletter
\let\NAT@parse\undefined
\makeatother
\usepackage[bookmarks=true]{hyperref} 

\input{macros}

\begin{document}

\title{Locally Optimal Estimation and Control of Cable Driven Parallel Robots using Time Varying Linear Quadratic Gaussian Control}

\author{
Gerry Chen, Seth Hutchinson, and Frank Dellaert%
\thanks{This material is based upon work supported by the National Science Foundation under Grant No. 2008302}
\thanks{Institute for Robotics and Intelligent Machines, College of Computing,
Georgia Institute of Technology,
Atlanta, GA, \{gchen328, seth, fd27\}@gatech.edu}
}

\maketitle

\begin{abstract}
\input{0_abstract}
\end{abstract}


\section{Introduction}\label{sec:introduction}
\input{1_intro}


\section{Approach}\label{sec:approach}
\input{2_approach}

\section{Experiments}\label{sec:experiments}
\input{3-experiments}

\section{Discussion}\label{sec:discussion}
\input{4-discussion}

\section{Conclusions and Future Work}\label{sec:conclusions}
\input{5-conclusions}

\appendix[Parameter Calibration and Tuning]
\input{6_appendix}


\bibliographystyle{IEEEtran}
\bibliography{bibs/PersonalLiterature,bibs/gerry}

\end{document}

%% file: macros.tex
\newcommand{\xzero}{\prescript{0}{}{x}}
\newcommand{\xd}{\prescript{d}{}{x}}
\newcommand{\um}{u_{mid}}
\newcommand{\xerror}{\tilde{x}}
\newcommand{\uerror}{\tilde{u}}
\newcommand{\umin}{u_{min}}
\newcommand{\umax}{u_{max}}
\newcommand{\xff}{x^*}
\newcommand{\uff}{u^*}
\newcommand{\zff}{z^*}
\newcommand{\deltax}{\delta x}
\newcommand{\deltau}{\delta u}
\newcommand{\deltaz}{\delta z}
\renewcommand{\l}{\bm{l}}
\newcommand{\ldot}{\dot{\bm{l}}}
\newcommand{\cov}{\Sigma}
\newcommand{\covdyn}{\prescript{d}{}{\cov}}
\newcommand{\covmeas}{\prescript{z}{}{\cov}}
\newcommand{\xhat}{\hat{x}}
\newcommand{\covpred}[1]{\cov_{#1| #1-1}}
\newcommand{\xhatpred}[1]{\xhat_{#1| #1-1}}
\newcommand{\covupdate}[1]{\cov_{#1}}
\newcommand{\xhatupdate}[1]{\xhat_{#1}}
\newcommand{\sqrtinfo}[1]{#1^{-\nicefrac{1}{2}}}

\newcommand{\sqrtinfomeas}{\sqrtinfo{\covmeas}}
\newcommand{\sqrtinfopred}[1]{\sqrtinfo{\covpred{#1}}}

\newcommand{\Kx}{\prescript{x}{}{K}}
\newcommand{\Ku}{\prescript{u}{}{K}}
\newcommand{\Kz}{\prescript{z}{}{K}}

\newcommand{\wrenchmatrix}{\bm{W}}

\DeclareMathOperator{\diag}{diag}

\makeatletter
\renewcommand*\env@matrix[1][*\c@MaxMatrixCols c]{%
  \hskip -\arraycolsep
  \let\@ifnextchar\new@ifnextchar
  \array{#1}}
\makeatother

%% file: 0_abstract.tex
We present a locally optimal tracking controller for Cable Driven Parallel Robot (CDPR) control based on a time-varying Linear Quadratic Gaussian (TV-LQG) controller.
In contrast to many methods which use fixed feedback gains, our time-varying controller computes the optimal gains depending on the location in the workspace and the future trajectory.  Meanwhile, we rely heavily on offline computation to reduce the burden of online implementation and feasibility checking. 
Following the growing popularity of probabilistic graphical models for optimal control, we use factor graphs as a tool to formulate our controller for their efficiency, intuitiveness, and modularity.
The topology of a factor graph encodes the relevant structural properties of equations in a way that facilitates insight and efficient computation using sparse linear algebra solvers.
%
We first use factor graph optimization to compute a nominal trajectory, 
then linearize the graph and apply variable elimination to compute the locally optimal, time varying linear feedback gains.
Next, we leverage the factor graph formulation to compute the locally optimal, time-varying Kalman Filter gains, and 
finally combine the locally optimal linear control and estimation laws to form a TV-LQG controller.
%
%
We compare the tracking accuracy of our TV-LQG controller to a state-of-the-art dual-space feed-forward controller on a 2.9m\,x\,2.3m, 4-cable planar robot and demonstrate improved tracking accuracies of \SI{0.8}{\degree} and \SI{11.6}{\mm} root mean square error in rotation and translation respectively.



%% file: 1_intro.tex
Given that cable robots have many desirable qualities, developing cable driven parallel robot (CDPR) tracking controllers that are accurate, robust, and easy to implement remains an important problem.
CDPRs consist of a set of cables, through which forces are transmitted, connected to (usually) fixed winches on one end and to the floating end effector on the other.  Due to the usage of cables to transmit forces, CDPR naturally scale well to large sizes \cite{Nan11ijmp_FAST_telescope,Saptarshi20nasa_lunartelescope,Miermeister16iros_CableRobotSimulator}, fast speeds/accelerations \cite{Chen22icra_GTGraffiti,Muller15cdpr_segesta_robot,Rajnoha17youtube_cable_endy}, and heavy loads \cite{Shahmiri16_cable_survey}.  Simultaneously, the inherent flexibility of cables makes achieving stiffness and accuracy challenging thereby limiting the utility of CDPR.
Finally, despite the remarkable progress by a vibrant community, CDPR remain largely limited to academic settings, with the notable exception of \textit{suspended} cable robots \cite{Cone85geo_skycam,Tang13_cable_telescope} which pose a distinct set of challenges due to their not being redundantly actuated.
We argue that the contrasting success of suspended cable robots compared with CDPR evidences that the complexity and difficulty in implementing CDPR tracking controllers is at least a large factor in curbing their adoption.

Tracking control for CDPRs has been well studied since at least the late 1980's and has progressed from simple joint-space methods to advanced nonlinear, adaptive, and robust methods. 
The earliest examples of CDPR tracking control commanded position-controlled servos to the cable lengths computed by inverse kinematics, but these require tedious, critical calibration and are sensitive to modelling errors \cite{Campbell95aero_charlotte_nasa_cdpr}. 
More advanced model-based techniques include operational-space PID, feedback linearization, sliding-mode control (SMC), linear quadratic (LQR and LQG), and MPC, among others \cite{Khosravi14tro_singular_perturbation}.  
Linear approaches (e.g. PID) with feedforward terms to handle most of the nonlinearity are perhaps the most common \cite{Lamaury13ICRA_CableControlFF, Gouttefarde15TRO_CableControl,Shang19tie_synchronous,Shang20mmt_synchronous}, but the use of static gains can make them sensitive to operating point, leading to non-uniform performance across the workspace and lack of foresight when approaching control limits \cite{Santos20thesis_mpc}.
Feedback linearization approaches have seen success in \emph{suspended} cable robot designs but require special considerations for redundantly actuated CDPR \cite{Santos20thesis_mpc,Begey19tro_position_controlled,Korayem17jirs_FeedbackLinearization_LQG,Korayem10ijrs_robust_feedback_linearization}.  
    Feedback linearization may also face difficulties modeling control limits.
SMC \cite{Slotine84ijc_smc} is perhaps most popular nonlinear approach for robust CDPR control, but is prone to chattering \cite{Niu17icorr_smc_chatters_medical} due to control discontinuities.  Although reducing chattering issues have been studied by a number of works \cite{Schenk15icstcc_smc_nonchattering,Schenk18cdpr_smc_nonchattering_realrobot,Zeinali10idetc_smc_chattering_free,Wang19isa_smc_AST,Picard20ifac_smc_linearfeedback}, chattering does not appear to be completely resolved and these methods rely on adaptive algorithms to fallback onto linear controllers which, again, may not be workspace-aware.
LQR and LQG have been primarily applied as tuning techniques for other controllers (e.g. tuning SMC \cite{Korayem12ijars_smc_lqr} or feedback linearization gains \cite{Korayem17jirs_FeedbackLinearization_LQG}) and therefore inherit their limitations.  Furthermore, applying LQR and LQG to systems that have been ``pre-linearized'' by other approaches tend to use fixed gains obtained using the infinite horizon LQR solution \cite{Korayem12ijars_smc_lqr, Lambert07tmech_blimp_cable_positioning}.
Linear \cite{Santos19cdpr_mpc_linear} and Nonlinear MPC \cite{Santos22tro_NMPC_tracking,Song22tro_mpc,Katliar17ifac_nmpc} are highly capable in providing consistent and robust performance near or even outside workspace boundaries \cite{Santos20thesis_mpc}, but MPC poses implementation challenges and must often make convexifying approximations to guarantee online convergence \cite{Santos20thesis_mpc}.

In this work, we propose a controller which is simple to implement but exhibits consistently accurate tracking performance across the workspace.
We achieve this by constructing a factor graph describing the kinematics/dynamics, objectives, and/or stochastic elements of the CDPR tracking control problem.  The factor graph is used to 
solve (offline) for an optimal trajectory, locally optimal linear time-varying (LTV) feedback control gains, and locally optimal LTV estimator.  We combine the control and estimator gains into a TV-LQG controller and demonstrate on a real robot.

\iffalse
Our contributions include the following:
\pagebreak
\begin{enumerate}
    \item describe a factor graph representing the kinematics/dynamics, objectives, and/or stochastic elements of the CDPR tracking control problem;
    \item show how the optimal nominal trajectory can be computed as the factor graph solution (\ref{ssec:approach_trajgen}); \raggedright
    \item show how the locally optimal linear time-varying estimator (\ref{ssec:approach_estimation}) and controller (\ref{ssec:approach_control}) can both be computed offline using factor graph elimination; and
    \item combine the nominal trajectory, estimator, and controller (\ref{ssec:approach_combine}) to demonstrate our method on a real robot (\ref{sec:experiments}).
    \rightskip=0pt
    \spaceskip=0pt
    \xspaceskip=0pt
    \relax
\end{enumerate}
%
%
\else\iffalse
Our contribution is to show how a factor graph representing the kinematics/dynamics, objectives, and/or stochastic elements of the CDPR tracking control problem is constructed and can be used to compute (offline) the:
\begin{enumerate}
    \item optimal nominal trajectory (\ref{ssec:approach_trajgen}); \raggedright
    \item locally optimal linear time-varying estimator (\ref{ssec:approach_estimation}); and 
    \item locally optimal linear time-varying controller (\ref{ssec:approach_control})
    \rightskip=0pt
    \spaceskip=0pt
    \xspaceskip=0pt
    \relax
\end{enumerate}
which are combined (\ref{ssec:approach_combine}) to demonstrate our method on a real robot (\ref{sec:experiments}).
\fi
\fi

%% file: 2_approach.tex
\begin{figure}
  \centering
  \includegraphics[page=1, trim=0 5.08in 7.28in+3pt 0, clip, width=0.8\linewidth]{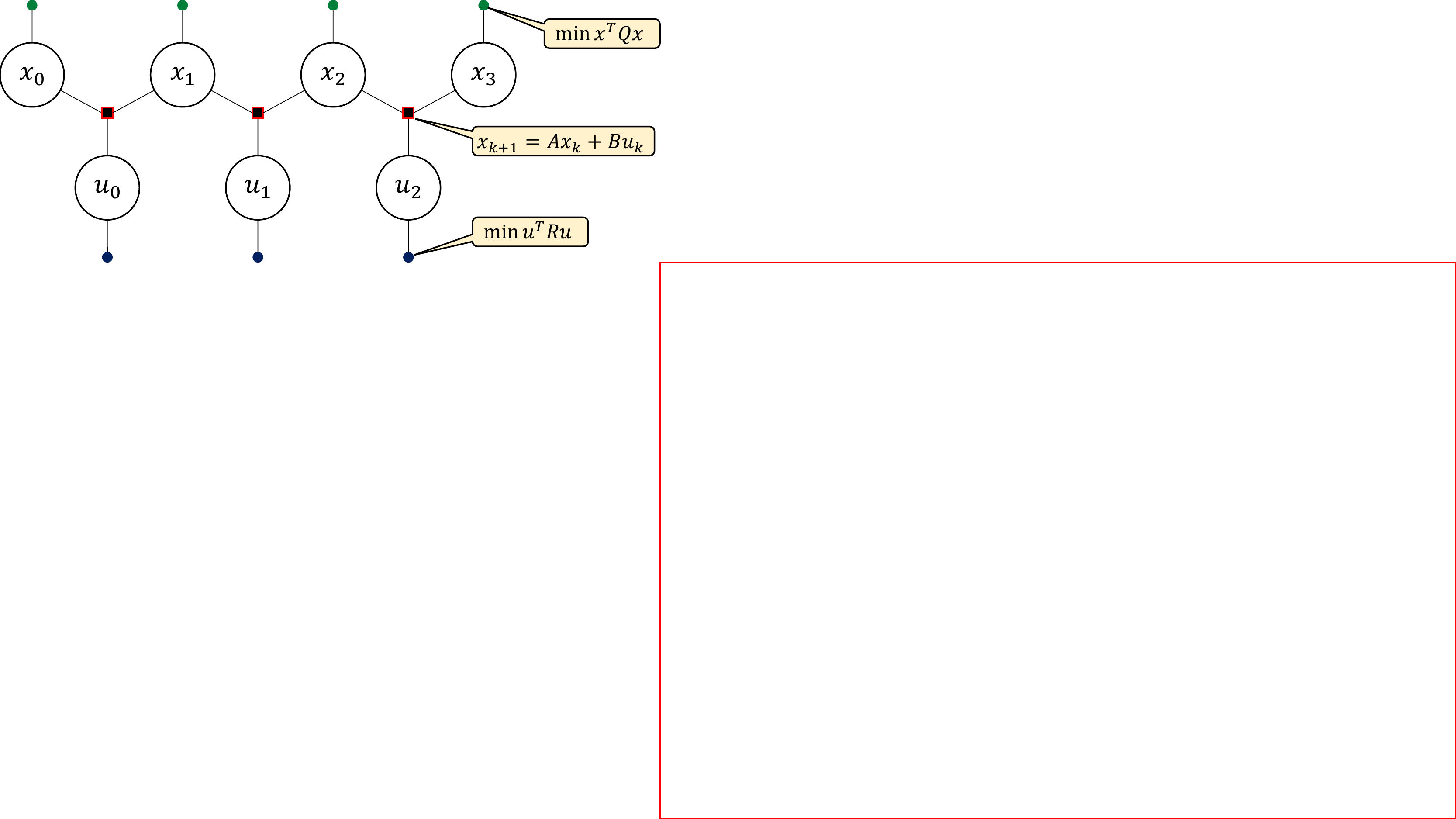}
  \caption{Prototypical example of a factor graph describing the LQR problem with 4 timesteps, adapted with permission from \cite{Chen19blog_lqr-blogpost}.  Large circles are variables and small black circles/squares are factors. $x_k$ represent state variables, $u_k$ represent control variables, black squares represent dynamics constraint factors, and black circles represent optimal control objectives.}
  \label{fig:standard_lqr}
\end{figure}

We frame our approach around factor graphs because we believe them to be useful abstractions for describing trajectory optimization and controls topics.  
Simultaneously, factor graphs are highly efficient, visually insightful, and modular, allowing efficient incorporation of nearly any sensor, control objective, and/or constraint with minimal modification \cite{Yang21icra_ecLQR}.
Probabilistic graphical models, including factor graphs, have seen success in state estimation and perception \cite{Dellaert17fnt_factorgraphsforrobotperception} and are becoming increasingly popular as tools for optimal control \cite{Dellaert21review_factorGraphsRobotics,Chen19blog_lqr-blogpost,Yang21icra_ecLQR}.
We will attempt to describe the most pertinent concepts about factor graphs as we go, but additional introductions to factor graphs can be found in \cite{Dellaert17fnt_factorgraphsforrobotperception,Koller09book_pgm_Koller_Friedman}.

In short, factor graphs are a graphical way to describe optimization problems where variables (unknowns to solve for) and factors (optimization objectives or constraints) are connected by an edge if the variable is involved in the objective/constraint.  Fig. \ref{fig:standard_lqr} shows a prototypical example of a factor graph describing the LQR problem.

\begin{figure}
  \centering
  \includegraphics[page=19, trim=0 5.25in 2.78in 0, clip, width=\linewidth]{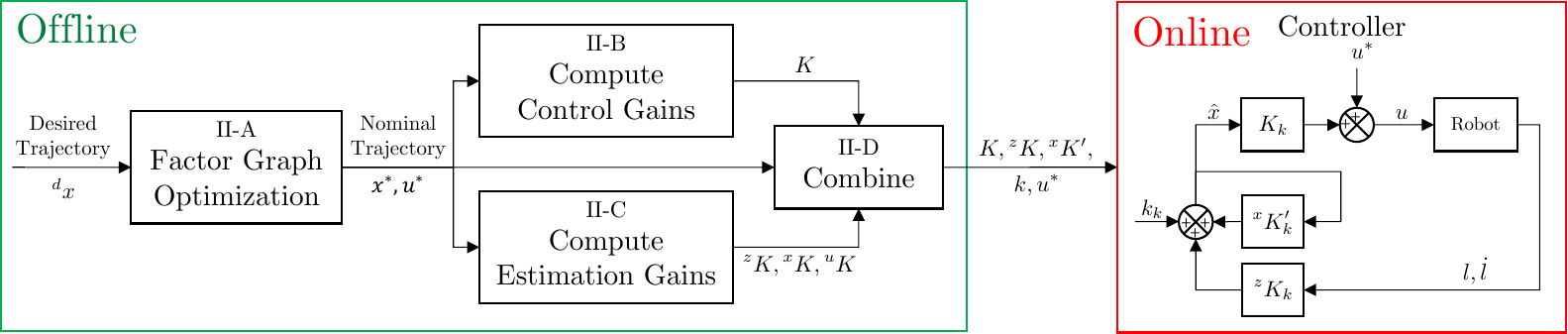}
  \caption{Our proposed controller leverages offline computations to simplify the online controller, which is locally optimal while requiring only 3 vector-matrix multiplications and 3 vector additions per update.  The offline stage, outlined in green, 
  generates time-varying gain matrices and feedforward vectors.
  The online stage, outlined in red, is a standard TV-LQG controller.}
  \label{fig:overview}
\end{figure}

Our algorithm, summarized in Fig. \ref{fig:overview}, consists of three offline stages followed by an online controller:
\begin{enumerate}[label=\Alph*.]
  \item Pre-compute a \textbf{nominal trajectory} that tracks the reference trajectory by ``solving'' a factor graph.  This is equivalent to trajectory generation using the iterative Linear Quadratic Regulator (iLQR) algorithm \cite{Li04icar_iLQR}.
  \item Pre-compute \textbf{time-varying LQR gain matrices} by \emph{linearizing} the graph around the nominal trajectory and using \emph{variable elimination} on the linearized graph.  Linearizing the graph is equivalent to applying a first-order Taylor expansion to convert the constrained \emph{nonlinear} least squares problem into a constrained \emph{linear} least squares problem (finite horizon LQR problem), and variable elimination is equivalent to using backward-induction to solve the Bellman equation.
  \item Pre-compute \textbf{time-varying Kalman Filter (KF) gain matrices} by creating a graph with stochastic factors instead of cost/constraint factors, \emph{linearizing} the graph around the nominal trajectory, and marginalizing at each time step.  The stochastic factors represent noise in the dynamics and/or measurements, and marginalizing at each time step is the Markov assumption the KF makes.
\end{enumerate}
Finally, we combine the LQR and KF gain matrices to form an online controller in the form of a TV-LQG controller.

\subsection{Trajectory Generation} \label{ssec:approach_trajgen}
\input{2-1_trajgen}

\subsection{Control} \label{ssec:approach_control}
\input{2-3_control}

\subsection{Estimation} \label{ssec:approach_estimation}
\input{2-2_estimation}

\subsection{Combined Optimal Feedback Controller} \label{ssec:approach_combine}
\input{2-4_combined}

%% file: 2-1_trajgen.tex

\begin{figure*}
  \centering
  \input{figs/ilqr_graph.tikz}
  \vspace*{-1.2em}
  \caption{Factor graph describing the full cable robot kinematics, dynamics, initial state, and state/control objectives, together with the equivalent optimization problem description (lower right).  The kinematics and dynamics constraints, which are represented by square factors and make up \eqref{eq:ilqr_dynamics}: $x_{k+1} = f_k(x_k, u_k)$, are defined according to standard cable robot literature.  For example, the orange cable tension factor encodes the constraint $\wrenchmatrix t=\mathcal{F}$ where $\wrenchmatrix$ is the wrench matrix computed from the pose $T$.
  Example references are given in the legend. $T$ denotes the pose of the end effector; $\mathcal{V}$ denotes the twist of the end effector; $\dot{\mathcal{V}}$ denotes the twist acceleration of the end effector; $l_i, \dot{l}_i, \ddot{l}_i, t_i$ denote the length, velocity, acceleration, and tension of cable $i$ respectively; $\tau_i$ denotes the torque commanded for motor $i$; and $\mathcal{F}_i$ denotes the wrench applied to the end effector by cable $i$.  Time indices are omitted for notational clarity.  Together, $T, \mathcal{V}$ make up the state $x$ and the torques $\tau_0, \ldots, \tau_3$ make up the control $u$.}
  \label{fig:ilqr_problem}
\end{figure*}
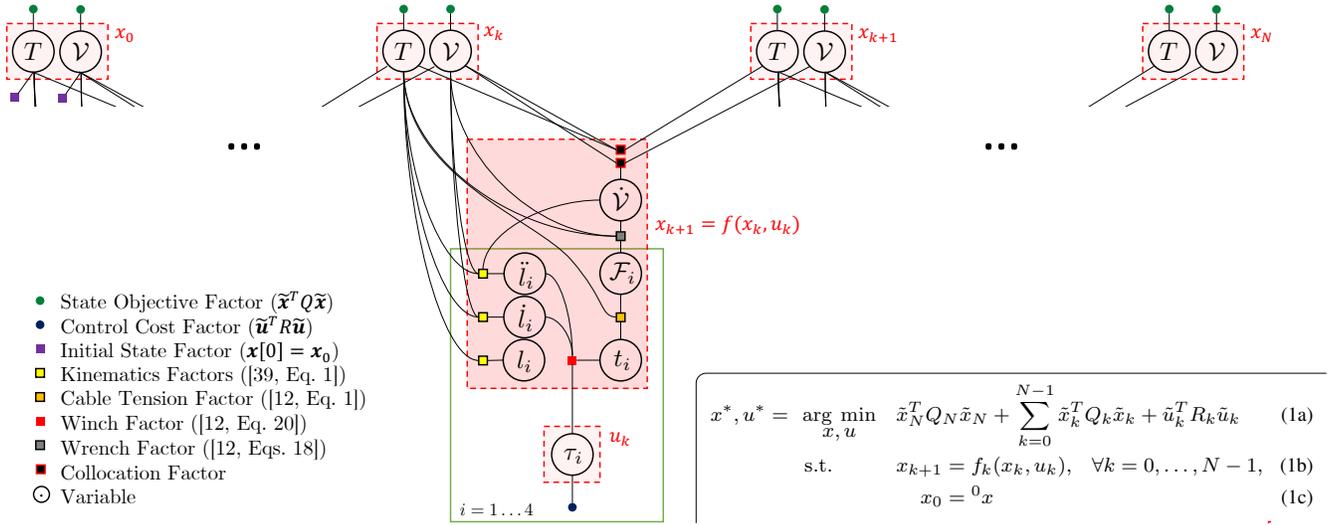

We seek to track the reference trajectory while also maintaining realizable control inputs.
Least-squares objectives are used to minimize both the tracking error, $\xerror \coloneqq x-\xd$, and the control input objective, $\uerror \coloneqq u-\um$, where $\xd$ is the desired trajectory and $\um \coloneqq \frac{1}{2}(\umin + \umax)$ as described by \cite{Pott09ckin_forcedist_closedform} as an approximation for maximizing the margin to control limits.
Constraints are given by the system dynamics.
Our trajectory generation problem can then be expressed as in Fig. \ref{fig:ilqr_problem}, which graphically depicts the trajectory generation problem as both a factor graph and the equivalent constrained nonlinear least squares problem: eq. \eqref{eq:ilqr_problem}.  The pose and twist, $T, \mathcal{V}$, make up $x$ and the tensions, $\tau_0, \tau_1, \tau_2, \tau_3$, make up $u$. 



We use the software library Georgia Tech Smoothing and Mapping (GTSAM) \cite{gtsam} to create and ``solve'' our factor graph, where ``solve'' is used to mean computing the solution to the equivalent constrained least squares problem.  We use the Levenberg-Marquardt optimizer with variable elimination for the inner-loop linear solver (default solver in GTSAM).  It can be shown that Levenberg-Marquardt with variable elimination is identical to iLQR as described by \cite{Li04icar_iLQR}, although we run this optimization offline.  As a result, feasibility (incl. control limits) and optimality can be checked prior to executing on the robot.

The computed solution to the factor graph is the nominal, feedforward trajectory and denoted $\xff, \uff$.

%% file: figs/ilqr_graph.tikz

\resizebox*{\linewidth}{!}{
\begin{tikzpicture}[      
    every node/.style={anchor=south west,inner sep=0pt,outer sep=0pt},
    x=1mm, y=1mm, inner sep=0pt, outer sep=0pt
  ]   %
  \node (fig1) at (8,0)
    {%
      \includegraphics[width=\linewidth, page=4, trim=0 2.55in 1.25in 0, clip]{figs/figs.pdf}
    };
  \node [text width=8.5cm, open rectangle sides=nw, rounded corners, ultra thin, draw, shape=open rectangle, inner sep=5pt] (fig3) at (105, 0) {%
    \footnotesize%
    \vspace*{-1em}
    \begin{argmini!}|s|
        {x, u}
        {\xerror_N^T Q_N \xerror_N + \sum_{k=0}^{N-1} \xerror_k^T Q_k \xerror_k + \uerror_k^T R_k \uerror_k \label{eq:ilqr_objective}}
        {\label{eq:ilqr_problem}}
        {\hspace*{-0.75em}\xff, \uff=}
        \addConstraint{x_{k+1}}{ = f_k(x_k, u_k),\quad\label{eq:ilqr_dynamics}}{\forall k=0,\ldots,N-1}
        \addConstraint{x_0}{= \xzero\label{eq:ilqr_init}}{}
    \end{argmini!}
  };
\end{tikzpicture}
}

%% file: 2-3_control.tex
We use a time varying linear-quadratic regular (LQR) for our controller which is computed by linearizing the system around $\xff, \uff$ and applying variable elimination.



\subsubsection{Linearizing the graph}
The linearization process is automatically handled by GTSAM, but mathematically it equates to applying a first-order Taylor expansion around $\xff, \uff$ to convert the general optimal control problem \eqref{eq:ilqr_problem} (which is transcribed as a constrained \emph{nonlinear} least squares optimization) into the LQR problem \eqref{eq:lqr_problem} (which is transcribed as a constrained \emph{linear} least squares problem).
Upon linearizing, we will define $\deltax_k \coloneqq x_k-\xff_k$ and $\deltau_k \coloneqq u_k-\uff_k$.





The result of linearizing is a graph with the same structure as in Fig. \ref{fig:ilqr_problem} except with all the factors representing either linear constraints or quadratic objectives.  Mathematically, the linearized graph now represents a \emph{linear} least squares problem (with linear equality constraints) which can be efficiently and exactly solved using variable elimination (which is equivalent to standard sparse linear algebra algorithms).

\subsubsection{Variable Elimination}


We first give a brief introduction to the variable elimination algorithm, then show how it can be applied to extract LQR gains from our linearized graph.

The \emph{variable elimination algorithm} is nothing more than an algorithmic method of solving for a variable and substituting the solution expression back into the system of equations.
Although GTSAM handles elimination for us, we can briefly summarize its underlying computation for linear problems as performing the block QR factorization on a subsystem:
$$
\begin{bmatrix}[cc|c] M_{11} & M_{12} & b_1 \\ M_{21} & M_{22} & b_2 \end{bmatrix} \begin{bmatrix} y \\ z \end{bmatrix}
=
Q\begin{bmatrix}[cc|c] R_{11} & S_{12} & d_1 \\ 0 & A_z & d_2 \end{bmatrix} \begin{bmatrix} y \\ z \end{bmatrix}
$$
where $M, b$ describes a linear least squares problem, $y$ are the variables we seek to eliminate, $z$ are the remaining variables (formally, $z$ is the \emph{separator}: set of variables not in $y$ that share a factor with any of the variables in $y$), and $Q,R,S,A,d$ are results of the factorization.  The top row encodes the solution $y = R_{11}^{-1}(d_1-S_{12}z)$ and the bottom row encodes the remaining system after substituting the solution for $y$ back in: $A_zz=d_2$.  In other words, factorization is equivalent to solving for a variable and substituting it back into the original system.
A comprehensive description of the variable elimination algorithm can be found in \cite{Dellaert17fnt_factorgraphsforrobotperception} for interested readers.

\begin{figure}
  \centering
  \resizebox*{\linewidth}{!}{\input{figs/lqr_graph.tikz}}
  \caption{After \emph{linearizing} Fig. \ref{fig:ilqr_problem} and applying \emph{variable elimination} to eliminate all the variables that are neither state nor control, we obtain this LQR graph (almost identical to Fig. \ref{fig:standard_lqr}), which can be equivalently represented as \eqref{eq:lqr_problem}.  Green and navy blue factors represent state and control costs respectively, red-outlined-black factors represent dynamics constraints, and the purple factor represents the initial state constraint.}
  \label{fig:linearized_graph}
\end{figure}
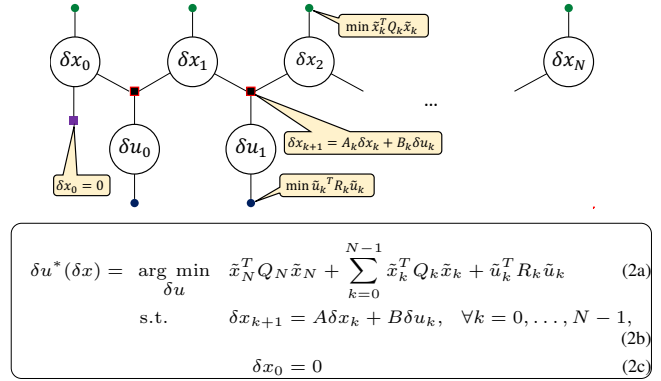

From the linearized graph, we first apply variable elimination to eliminate all variables except $\deltax, \deltau$, meaning we eliminate $y = \{\delta\l, \delta\ldot, \delta\ddot{\l}, \delta t, \delta\mathcal{F}, \delta\dot{\mathcal{V}}\}$ but leave $z = \{\delta T, \delta\mathcal{V}, \delta\tau\}$ (for every timestep).
Reiterating, this is equivalent to solving for $y$ first and substituting back into \eqref{eq:lqr_problem}.
After eliminating $y$, the resulting graph/system for $z$ will have the form of Fig. \ref{fig:linearized_graph} which can be equivalently described by the linear least squares problem \eqref{eq:lqr_problem}.  Also note that the computer has performed all the algebraic manipulation for us.  Introducing e.g. additional parameters, periodicity constraint factors, and state-dependent control limits are all possible with minimal designer effort.

Finally, we can also use variable elimination to derive the well known result of linear-quadratic control \cite{Todorov06chapter_optimal_control} that the solution $\deltau^*(\deltax)$ can be expressed as the linear control law:
\begin{align}
  \deltau_k^*(x_k) = K_k\deltax_k. 
  \label{eq:LQR}
\end{align}
Notice that, since $\xff, \uff$ satisfy \eqref{eq:ilqr_problem}, they are also guaranteed to satisfy \eqref{eq:lqr_problem}.  Thus when $\deltax_k=0$, $\deltau_k^*(0)=0$ so the affine term is zero.

\begin{figure}
  \centering
  \scriptsize
  \parbox[t]{0.85\linewidth}{\includegraphics[page=8,width=\linewidth,trim=0 4.72in 6.95in 0, clip]{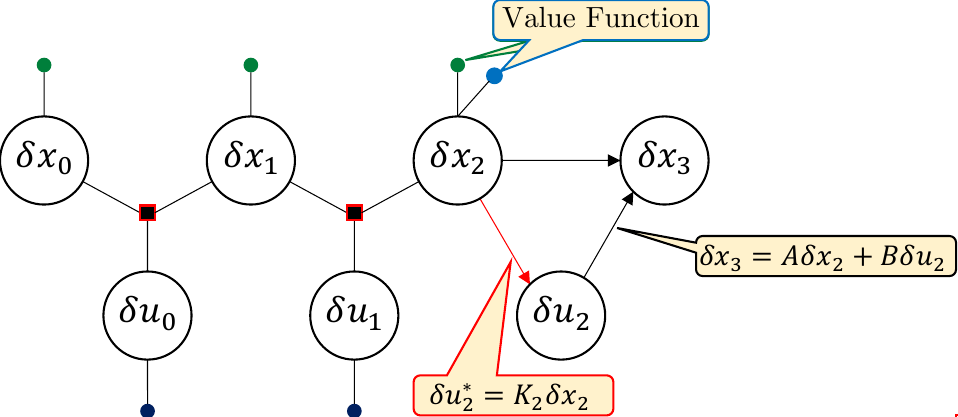}}\\~\\
  \caption{Variable elimination is used to compute the optimal time-varying feedback control law starting at the final timestep and eliminating back to the beginning.  This diagram depicts eliminating two variables: $\deltax_3, \deltau_3$.  The blue factor is the cost that gets propogated upon ``substituting'' the optimal value back into the system, which makes up the value function.}
  \label{fig:lqr_elimination}
\end{figure}

To obtain the LQR gains, $K_k$, from the linear graph in Fig. \ref{fig:linearized_graph},
we eliminate the states and controls one at a time in the order $\deltax_N, \deltau_{N-1}, \deltax_{N-1}, \ldots, \deltau_0, \deltax_0$.
It was proved in \cite{Chen19blog_lqr-blogpost,Yang21icra_ecLQR} that this variable elimination process produces the finite horizon discrete LQR solution.
As a brief intuition for the proof, this elimination follows the standard logic (backward induction on the Bellman equation) for deriving the finite horizon discrete algebraic ricatti equation: given the value function for $x_{k+1}$, solve for the $u_k$ which minimizes the value function then substitute $u_k^*$ back into the value function to obtain the new value function for $x_k$.
Fig. \ref{fig:lqr_elimination} visually illustrates the elimination process.


%% file: figs/lqr_graph.tikz
\begin{tikzpicture}[      
    every node/.style={anchor=south east,inner sep=0pt,outer sep=0pt},
    x=1mm, y=1mm, inner sep=0pt, outer sep=0pt]
\node (img) at (10, 2) {   
    \includegraphics[page=5,width=.85\linewidth,trim=0 5.08in 6.90in 0,clip]{figs/figs.pdf}
    \hspace*{.075\linewidth}
};
\node [anchor=north east, text width=\linewidth, open rectangle sides=, rounded corners, ultra thin, draw, shape=open rectangle] (eq) at (10, 0) {
  \scriptsize
  \begin{argmini!}|s|
    {\deltau}
    {\xerror_N^T Q_N \xerror_N + \sum_{k=0}^{N-1} \xerror_k^T Q_k \xerror_k + \uerror_k^T R_k \uerror_k \label{eq:lqr_objective}}
    {\label{eq:lqr_problem}}
    {\deltau^*(\deltax) = }
    \addConstraint{\deltax_{k+1}}{ = A\deltax_k + B\deltau_k,\quad\label{eq:lqr_dynamics}}{\forall k=0,\ldots,N-1}
    \addConstraint{\deltax_0}{= 0\label{eq:lqr_init}}{}
  \end{argmini!}
};
\end{tikzpicture}

%% file: 2-2_estimation.tex
We can express our estimation problem by converting the factor graph in Fig. \ref{fig:ilqr_problem} to replace controls factors by estimation factors, followed by linearization and marginalization.

\subsubsection{Estimation Factor Graph}
To convert from the optimal control factor graph to the estimation factor graph, we make the following factor replacements:
\begin{center}
\begin{tabular}{c@{~$\longrightarrow$~}c}
  \fbox{\parbox{2.6cm}{\centering
    $\xerror_k^TQ_k\xerror_k$ \\
          (state cost)
  }} & \fbox{\parbox{4cm}{\centering
    $\norm{h_k(x_k)-z_k}_{\covmeas^{-1}_k}^2$ \\
        (stochastic measurement)
  }}\\[1.5em]
  \fbox{\parbox{2.6cm}{\centering
    $x_{k+1}=f(x_k, u_k)$ \\
      (dyn. constr.)
  }} & \fbox{\parbox{4cm}{\centering
    $\norm{x_{k+1} - f(x_k, u_k)}_{\covdyn^{-1}_k}^2$ \\
      (modelling uncertainty)
  }} \\[1.5em]
  \fbox{\parbox{2.6cm}{\centering
    $\uerror_k^TR_k\uerror_k$ \\
      (control cost)
  }} & \fbox{\parbox{4cm}{\centering
    $\norm{u_k - u_k^*}_{\covdyn'^{-1}_k}^2$ \\
      (input disturbance)
  }}
\end{tabular}
\end{center}
where $\norm{\square}_{\cov^{-1}} \coloneqq \square^T \cov^{-1}\square$ denotes the Mahalanobis norm; $z_k\coloneqq[\l_k, \ldot_k]$ denotes the measurement vector at timestep $k$; $u_k^*$ and $u_k$ denote the commanded and realized control torques vectors respectievly; $h(x_k)$ denotes the inverse kinematics to compute the cable length given the state; and $\covmeas, \covdyn, \covdyn'$ denote the measurement, dynamics, and input covariances respectively.

The factor graph now represents the system:
\begin{align}
  x_{k+1} &= f(x_k, u_k+v_k') + v_k,&v_k'\sim \mathcal{N}(0, \covdyn_k')\\
    & & v_k\sim \mathcal{N}(0, \covdyn_k)\\
  z_k &= h(x_k) + w_k & w_k\sim \mathcal{N}(0, \covmeas_k)
\end{align}

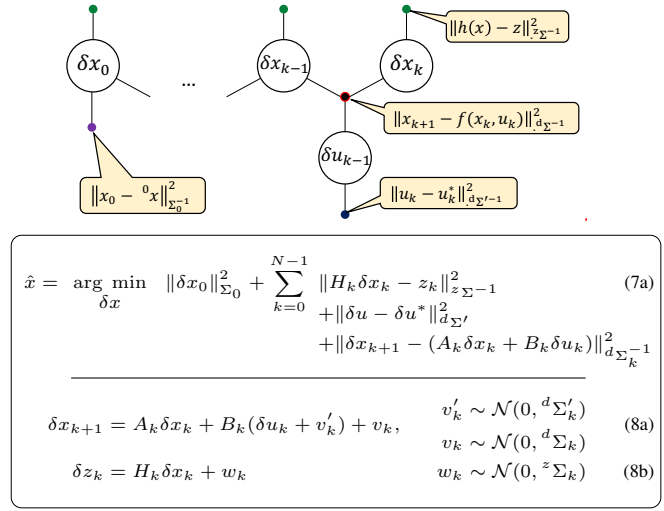
\begin{figure}
  \centering
  \input{figs/kf_graph.tikz}\vspace*{-1em}
  \caption{Nonlinear estimation factor graph (top) and the linearized estimation problem expressed as a minimization problem (middle) and system (bottom).  After linearizing, we introduce the variables $\deltaz_k \coloneqq z-\zff$ and $\zff \coloneqq h(\xff_k)$, and $H_k \coloneqq \left.\frac{\partial h_k}{\partial x_k}\right|_{\xff_k}$.}
  \label{fig:kalman_filter_graph}
\end{figure}

\subsubsection{Linearization}
Next, we linearize and eliminate the intermediate variables the same way as in Section \ref{ssec:approach_control} to obtain the factor graph shown in Fig. \ref{fig:kalman_filter_graph} which is equivalent to the system \eqref{eq:kf_system} in Fig. \ref{fig:kalman_filter_graph}.

After linearizing, $v_k'$ can be combined with $v_k$ by applying the transformation:
\begin{align}
  \deltax_{k+1} &= A_k\deltax_k + B_k(\deltau_k + v_k') \\
  &= A_k\deltax_k + B_k\deltau_k + B_kv_k'
\end{align}
where $v_k'\sim \mathcal{N}(0, \covdyn')$ so $v_k\sim \mathcal{N}(0, (B_k^T\covdyn'^{-1} B_k)^{-1})$ and $\covdyn_k = (B_k^T \covdyn'^{-1} B_k)^{-1}$.  This is useful because most disturbances in our CDPR setup come in the form of parasitic torques/tensions in the cable as opposed to external wrenches acting directly on the end effector.  Meanwhile, removing $v'_k$ simplifies the factor graph and the notation.


\subsubsection{Marginalization to Extract Kalman Gains}
We now seek a Kalman Filter to estimate $\xhat_k = E[\deltax_k]$.
Specifically, we would like to pre-compute the Kalman Gains for this filter to avoid needing to compute them online.  Although, to the best of our knowledge, factor graphs cannot directly extract the Kalman Gains the way we were able to extract the LQR gains, they can nevertheless get us most of the way and provide intuition for the final expression.

\begin{figure}
  \centering
  \includegraphics[page=13,width=\linewidth,trim=0 5.09in 3.38in 0,clip]{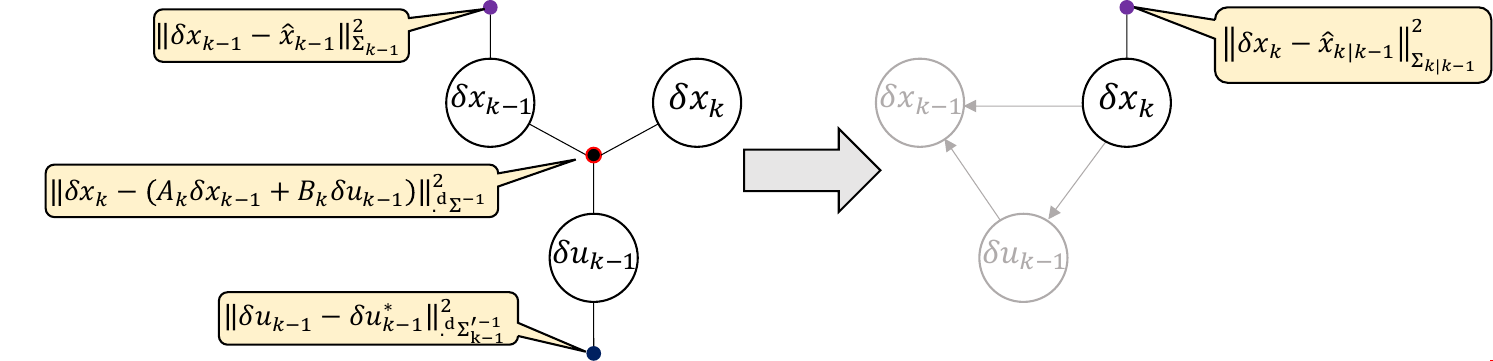}
  \caption{Kalman Filter ``Predict'' step eliminates $\deltax_{k-1}, \deltau_{k-1}$ to obtain $\covpred{k}$.}
  \label{fig:kf_predict}
\end{figure}

\begin{figure}
  \centering
  \includegraphics[page=16,width=0.85\linewidth,trim=0 6.12in 6.39in 0,clip]{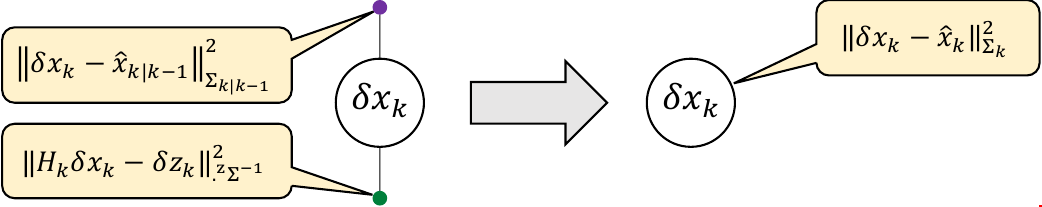}
  \caption{Kalman Filter ``Update'' step eliminates $\deltax_k$ to obtain $\covupdate{k}$.}
  \label{fig:kf_update}
\end{figure}

First, we can use marginalization to compute all the \textit{a priori} and \textit{a posteriori} covariances.  In these cases, marginalization can be computed by simply eliminating every variable other than the one we seek the marginal of.  Fig. \ref{fig:kf_predict} and \ref{fig:kf_update} show how variable elimination is used to compute the new covariances after predicting and updating respectively.  This computation is handled by GTSAM offline and $\cov_{1|0}, \covupdate{1},\ldots, \covpred{N}, \covupdate{N}$ are the only results we need for the next step.




We now obtain expressions to compute $\xhat$ given the covariance matrices we just computed.
The equation to predict $\xhatpred{k}$ is straightforward since all noises are zero-mean:
\begin{equation}
  \xhatpred{k} = A_k\xhatupdate{k-1} + B_k\deltau_{k-1}.
\end{equation}

The equation to update $\xhatupdate{k}$ is less obvious, but more intuitive by observing Fig. \ref{fig:kf_update}: the optimal estimate for $\deltax_k$ in Fig. \ref{fig:kf_update} is the one which makes the best compromise between $\norm{\deltax_k-\xhatpred{k}}_{\covpred{k}^{-1}}^2$ and $\norm{H_k\deltax_k - \deltaz_k}_{\covmeas_k^{-1}}^2$.  We expect the optimal choice of $\deltax_k$ to be a weighted mean of the two factors, and indeed with a bit of algebra it can be proven that it is:
\begin{align}
  \hspace*{-1em}\norm{\deltax_k - \xhatupdate{k}}_{\covupdate{k}^{-1}}^2 =&~
  \norm{\deltax_k-\xhatpred{k}}_{\covpred{k}^{-1}}^2 \nonumber \\[0.25em]
    &+ \norm{H_k\deltax_k - \deltaz_k}_{\covmeas_k^{-1}}^2 \nonumber \\[0.5em]
  \deltax_k^T \covupdate{k}^{-1}\xhatupdate{k} =&~ \deltax_k^T\covpred{k}^{-1}\xhatpred{k} + \deltax_k^TH_k^T\covmeas_k^{-1}\deltaz_k \nonumber \\[0.5em]
  \covupdate{k}^{-1}\xhatupdate{k} =&~ \covpred{k}^{-1}\xhatpred{k} + H_k^T\covmeas_k^{-1}\deltaz_k \nonumber \\[1em]
  \xhatupdate{k} =&~ \covupdate{k}\left( \covpred{k}^{-1} \xhatpred{k} + H_k^T\covmeas_k^{-1}\deltaz_k \right)
\end{align}

Finally,
the predict and update steps can be combined to form a single state estimator update equation:
\begin{align}
  \xhat_k =& \Kx_k \xhat_{k-1} 
        + \Ku_k \deltau_{k-1} 
        + \Kz_k \deltaz_k \label{eq:KF}
\end{align}
where $\Kx_k \coloneqq \covupdate{k}\sqrtinfopred{k}A_k,\quad \Ku_k \coloneqq \covupdate{k}\sqrtinfopred{k}B_k,$ and $\Kz_k \coloneqq \covupdate{k}H_k^T \sqrtinfomeas_k$ are pre-computed offline.

%% file: figs/kf_graph.tikz
\begin{tikzpicture}[      
    every node/.style={anchor=south east,inner sep=0pt,outer sep=0pt},
    x=1mm, y=1mm, inner sep=0pt, outer sep=0pt]
\node (img) at (10, 2) {   
  \includegraphics[page=11,width=.8\linewidth,trim=0 5.08in 7.50in 0, clip]{figs/figs.pdf}
  \hspace*{.1\linewidth}
  };
\node [anchor=north east, text width=\linewidth, open rectangle sides=, rounded corners, ultra thin, draw, shape=open rectangle] (eq) at (10, 0) {
  \parbox{0.99\linewidth}{
  \scriptsize
  \begin{argmini!}|s|
    {\deltax}
    {\norm{\deltax_0}_{\cov_0}^2 + \sum_{k=0}^{N-1}\hspace*{-.4em}
    \begin{array}[t]{l}       \label{eq:kf_objective}
      \norm{H_k\deltax_k-z_k}_{\covmeas^{-1}}^2 \\
      + \norm{\deltau - \deltau^*}_{\covdyn'}^2 \\
      + \norm{\deltax_{k+1}-(A_k\deltax_k+B_k\deltau_k)}_{\covdyn_k^{-1}}^2
    \end{array} \hspace*{-3em}
    }
    {\label{eq:kf_problem}}
    {\xhat = }
  \end{argmini!}\\[-2.5em]
  \centering\rule[-1em]{0.8\linewidth}{0.4pt}
  \begin{subequations}
  \begin{align}
    \deltax_{k+1} &= A_k\deltax_k + B_k(\deltau_k+v_k') + v_k,&\begin{aligned}v_k'\sim \mathcal{N}(0, \covdyn_k')\\
        v_k\sim \mathcal{N}(0, \covdyn_k)\end{aligned} \label{eq:kf_system_x}\\
    \deltaz_k &= H_k\deltax_k + w_k &w_k\sim \mathcal{N}(0, \covmeas_k) \label{kf_system_z}
  \end{align} \label{eq:kf_system}
  \end{subequations}
  }\hfill\\~\\[-1.5em]~
};
\end{tikzpicture}

%% file: 2-4_combined.tex
Finally, we can use the standard LQG result: the stochastic optimal control policy is to feed the optimal estimator into the optimal controller, for systems with only state- and control- independent noise.
From \eqref{eq:KF}, we obtain an optimal estimate $\xhat_k$ for $\deltax_k$ so LQG states that we may use $\xhat_k$ in place of $\delta_x$ in \eqref{eq:LQR}:
\begin{align}
  \deltau_k &= K_k\xhat_k \label{eq:lqg_lqr_part} \\
  \xhat_k &= \Kx_k \xhat_{k-1} + \Ku_k \deltau_{k-1} + \Kz_k \deltaz_k \label{eq:lqg_kf_part}
\end{align}

Making the assumption that our controller always follows the optimal controller, we can also substitute \eqref{eq:lqg_lqr_part} into \eqref{eq:lqg_kf_part}:
\begin{align}
  \xhat_k &= \Kx_k' \xhat_{k-1}  + \Kz_k \deltaz_k
\end{align}
where $\Kx_k' \coloneqq \Kx_k + \Ku_kK_k$.

And finally, for completeness, we can substitute our $\delta$ expressions to obtain our final time varying LQG controller:
\begin{align}
  \xhat_k &= \Kx_k' \xhat_{k-1} + \Kz_k z_k + k_k \label{eq:online_estimator} \\
  u_k &= K_k \xhat_k + \uff_k \label{eq:online_controller}
\end{align}
where $k_k \coloneqq - \Kz_k \zff_k$.

The only computations that need to be run online are \eqref{eq:online_estimator} and \eqref{eq:online_controller}, which require only 3 matrix multiplications and 3 vector additions per update.  Meanwhile, $\mathcal{K}_k \coloneqq {K_k, \Kx_k, \Kz_k, \Ku_k, \uff, k_k}$ are pre-computed offline.

\subsubsection{Interpolation}
It may be desirable to run the controller at a faster rate than the set of precomputed $\mathcal{K}$, since $\mathcal{K}$ may be only slowly changing while we typically want a controller update rate of at least several hundred Hz.  We can devise a simple ``zero-th'' order interpolation:
\begin{align}
  \xhat(k\,dt+\Delta t) = \Kx_k'\xhat_{k-1} + \Kz_k z(k\,dt+\Delta t) + k_k
\end{align}

Although we found in our experiments that this interpolation is sufficient, more intelligent interpolation schemes may be investigated.


%% file: 3-experiments.tex
\begin{figure}
  \centering
  \includegraphics[width=0.6\linewidth]{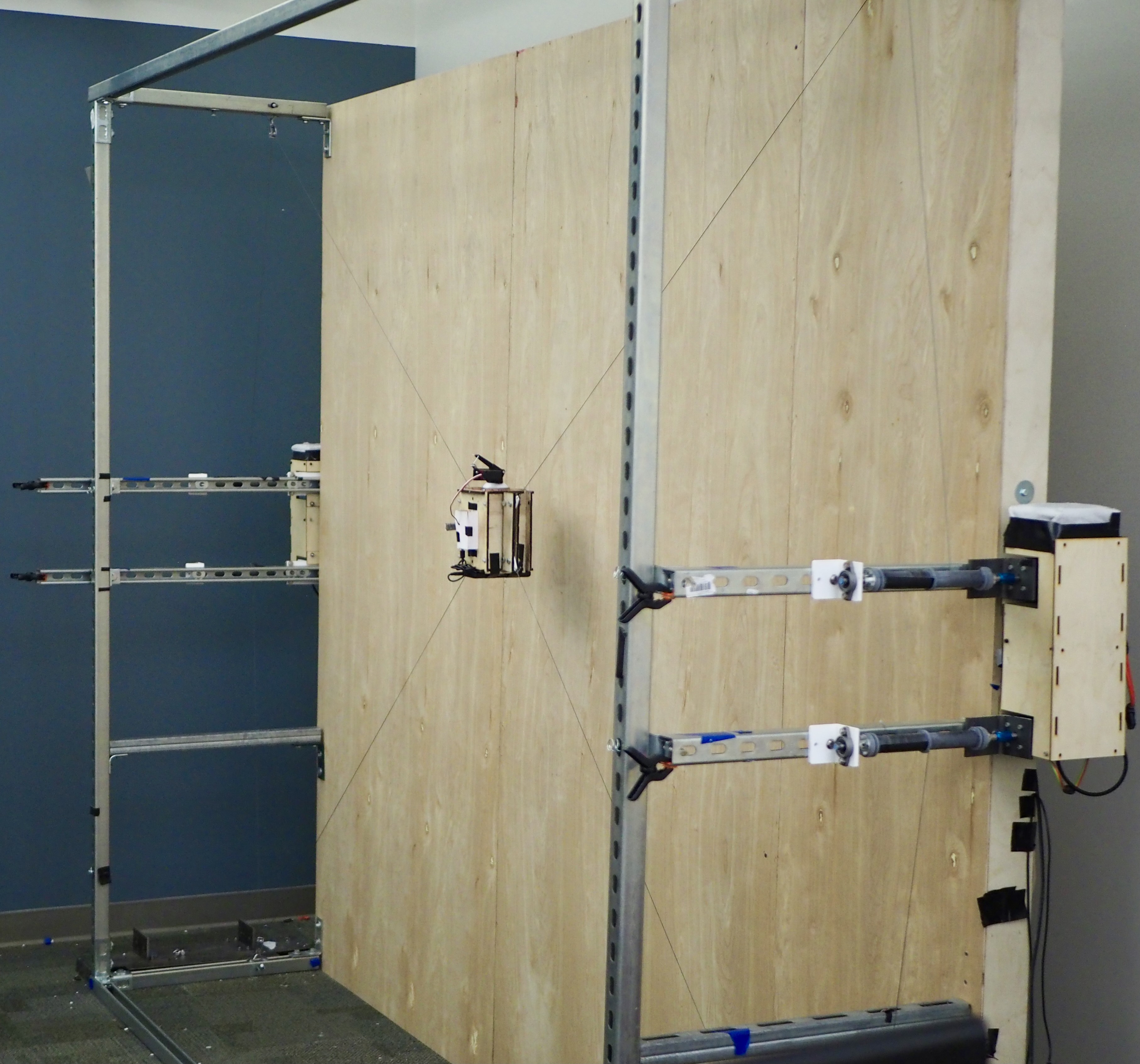}
  \caption{Image of the CDPR (from \cite{Chen22icra_GTGraffiti}) we use for experimental results.}
  \label{fig:photograph}
\end{figure}

We validate our approach experimentally by evaluating the trajectory tracking performance of our controller compared to a baseline controller implementation.  We use the planar SE(2), 4-cable redundantly actuated robot shown in Fig. \ref{fig:photograph}.  Our offline stage is discretized at 100Hz and the online controller runs at 1kHz (for both ours and the baseline).  We will give a brief description of the baseline controller then compare the tracking accuracies using a sample trajectory.

\subsection{Baseline Controller}
We model our baseline controller on the dual-space feedforward controller in \cite{Gouttefarde15TRO_CableControl} as a standard controller used and demonstrated to be successful in tracking control of (redundantly actuated) CDPR.  As compared to SMC-based and MPC-based controllers, PID-based controllers appear to be more widely used in redundantly actuated cable robots, of which \cite{Gouttefarde15TRO_CableControl} appears to be state of the art.

\begin{figure}
  \centering
  \includegraphics[page=6,width=\linewidth,trim=0 5.36in 3.14in 0, clip]{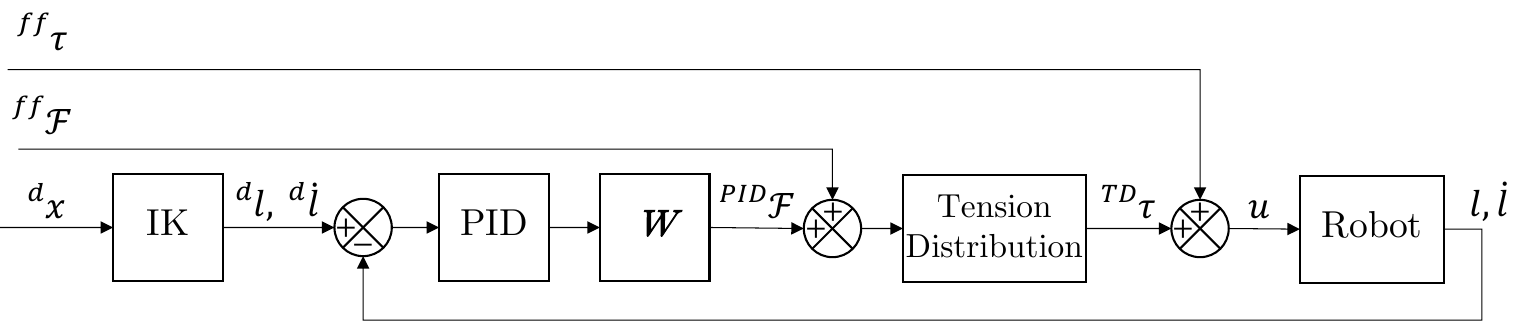}
  \caption{Baseline controller, adapted from \cite{Gouttefarde15TRO_CableControl}.}
  \label{fig:gouttefarde_controller}
\end{figure}

The baseline controller applies a PID controller in the joint-space, converts the correction to the operational-space to add inertial feedforward terms, applies a tension distribution algorithm to convert back into joint torques, and finally adds friction feedforward terms to generate the commanded torque.
The controller is summarized in Fig. \ref{fig:gouttefarde_controller}.

For the tension distribution algorithm, we apply some minor modifications since we have $n+1$ cables (4-cables for a 3DoF planar robot) whereas the tension distribution algorithm given in \cite{Gouttefarde15TRO_CableControl} is for $n+2$ cables.
The $n+1$ cable case is significantly easier to solve than the $n+2$ cable case, and it suffices to exhaustively check each of the 8 control limits (4 upper-bound, 4 lower-bound) to find the 1-dimensional polytope (line segment) of feasible configurations.
We then use \cite[IV.A]{Gouttefarde15TRO_CableControl}: 2-Norm Optimal Solution.  We were unable to successfully control the robot with the given objective $\norm{u}^2_2$, but based on \cite{Pott09ckin_forcedist_closedform}, using $\norm{u-\um}^2_2$ instead produced good performance. 

\subsection{Robot and Controller Parameters}
The robot and controller parameters are given in Tables \ref{tab:robot_parameters} and \ref{tab:controller_parameters}, respectively.  Details on the measurement, calibration, and tuning of values are provided in the Appendix.

\begin{table}
  \centering
  \caption{Geometric, Inertial, and Friction Parameters}
  \begin{tabular}[t]{rc|c}
    \multirow{4}{*}{\parbox{6em}{\centering Frame\\Mounting\\Location (m)}}
    & 0 & $(2.815, 0.000)$ \\
    & 1 & $(2.845, 2.239)$ \\
    & 2 & $(0.033, 2.225)$ \\
    & 3 & $(0.000, 0.000)$ \\ \hline
    \multirow{4}{*}{\parbox{6em}{\centering End Effector\\Mounting\\Location (m)}}
    & 0 & $(+0.063, -0.060)$ \\
    & 1 & $(+0.063, +0.060)$ \\
    & 2 & $(-0.063, +0.060)$ \\
    & 3 & $(-0.063, -0.060)$ \\ \hline
    End effector Inertia & $\mathcal{G}$ &
      $\diag\left(\left[\begin{array}{c}
          \SI{7.79}{\kg\mm\squared}\\
          \SI{0.727}{kg}\\
          \SI{0.727}{kg}
      \end{array}\right]\right)$\\[1.5em] 
    Winch inertia & & \SI{19.6}{\kg\mm\squared} \\ \hline
    Viscous friction & $F_v$ & \SI{0.002}{\N\m\s} \\
    Static friction & $F_s$ & \SI{0.12}{\N\m} \\
    $\tanh$ parameter & $\mu$ & \SI{0.19}{\s/\radian}
  \end{tabular} 
  \label{tab:robot_parameters}
\end{table}

\captionsetup[subtable]{position=top}
\begin{table}
  \caption{LQG and Baseline Controller Parameters}\vspace*{-2em}
  \hspace*{-1em}\subfloat[LQG Parameters]{
  \setlength{\tabcolsep}{3pt}
  \begin{tabular}[t]{c|c}
    $\cov_0$ & $\diag$([\SI{5.7}{\degree} \SI{0.1}{\m} \SI{0.1}{\m} 0 0 0]) \\
    $\covmeas$ & $\diag$([$(\SI{0.0018}{\m})$x$4~(\SI{0.04}{\m/\s})$x$4$]) \\
    $\covdyn'$ & $(\SI{0.059}{\N\m})I_{4x4}$ \\
    $Q$ & $\diag([10^2 ~ 10^4 ~ 10^4 ~ 0 ~ 0 ~ 0])$ \\
    $R$ & $I_{6x6}$
  \end{tabular}}
  \subfloat[Baseline Controller Gains]{
    \parbox[t]{3.3cm}{
    \centering
    \setlength{\tabcolsep}{3pt}
    \begin{tabular}[t]{c|l}
      $K_p$ & \SI{3e3}{\newton/\m} \\
      $K_i$ & \SI{5e3}{\newton/\m.\s}\\
      $K_d$ & \SI{1e1}{\newton.\s/\m}
    \end{tabular}
  }
  }
  \label{tab:controller_parameters}
\end{table}

\subsection{Tracking Accuracy}

To compare the tracking performances of the two controllers on a real robot, we use a trajectory composed of concentric diamonds in the center 1.5$\times$1m area of the workspace with a maximum speed and acceleration of 0.5m/s and 1m/s$^2$ respectively.  We use an 8-camera OptiTrack motion capture system (\SI{0.4}{\mm} accuracy) to measure the ground-truth end effector poses and assess the tracking accuracy.

\begin{figure}
  \centering
  \includegraphics[width=0.75\linewidth]{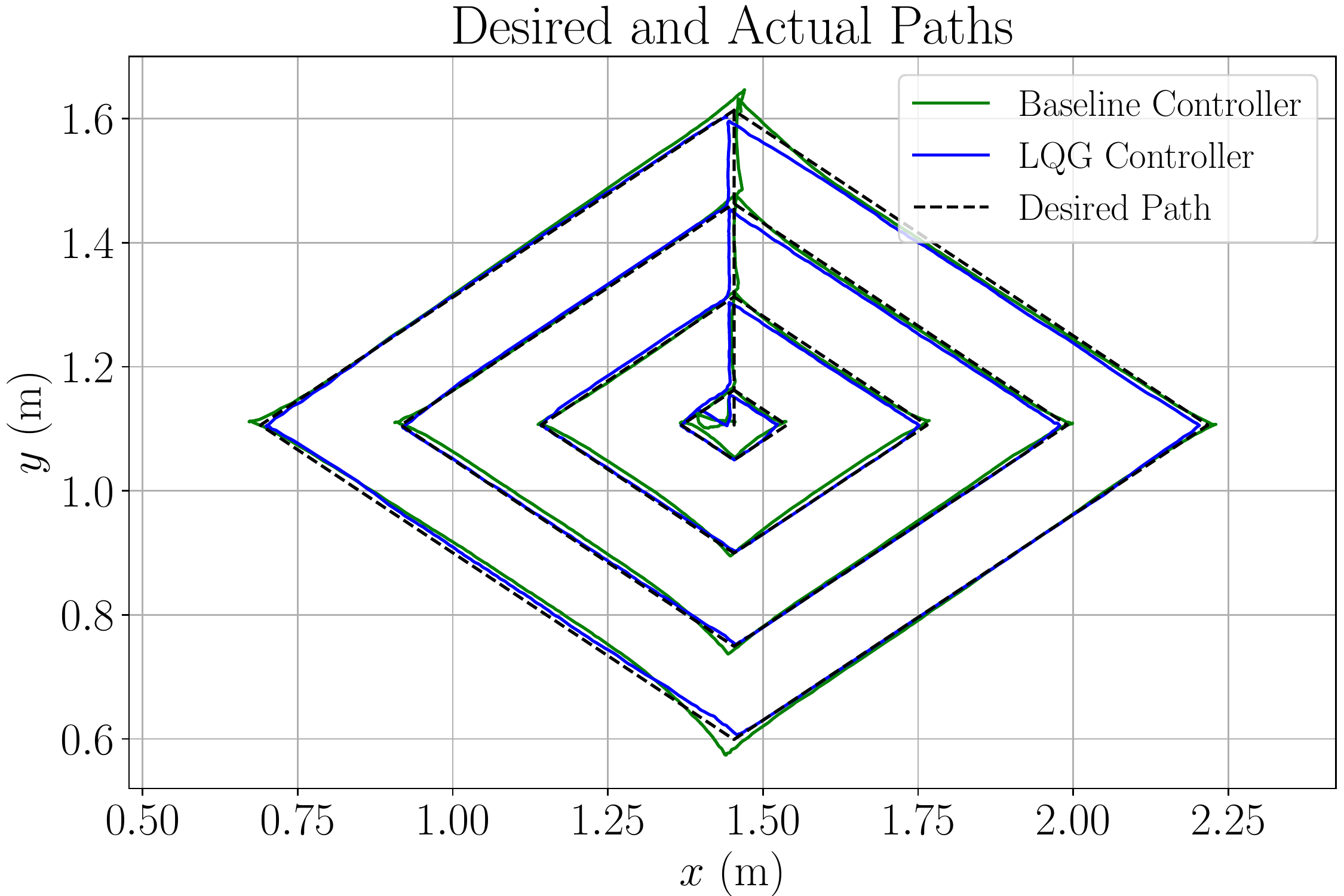}
  \includegraphics[width=\linewidth]{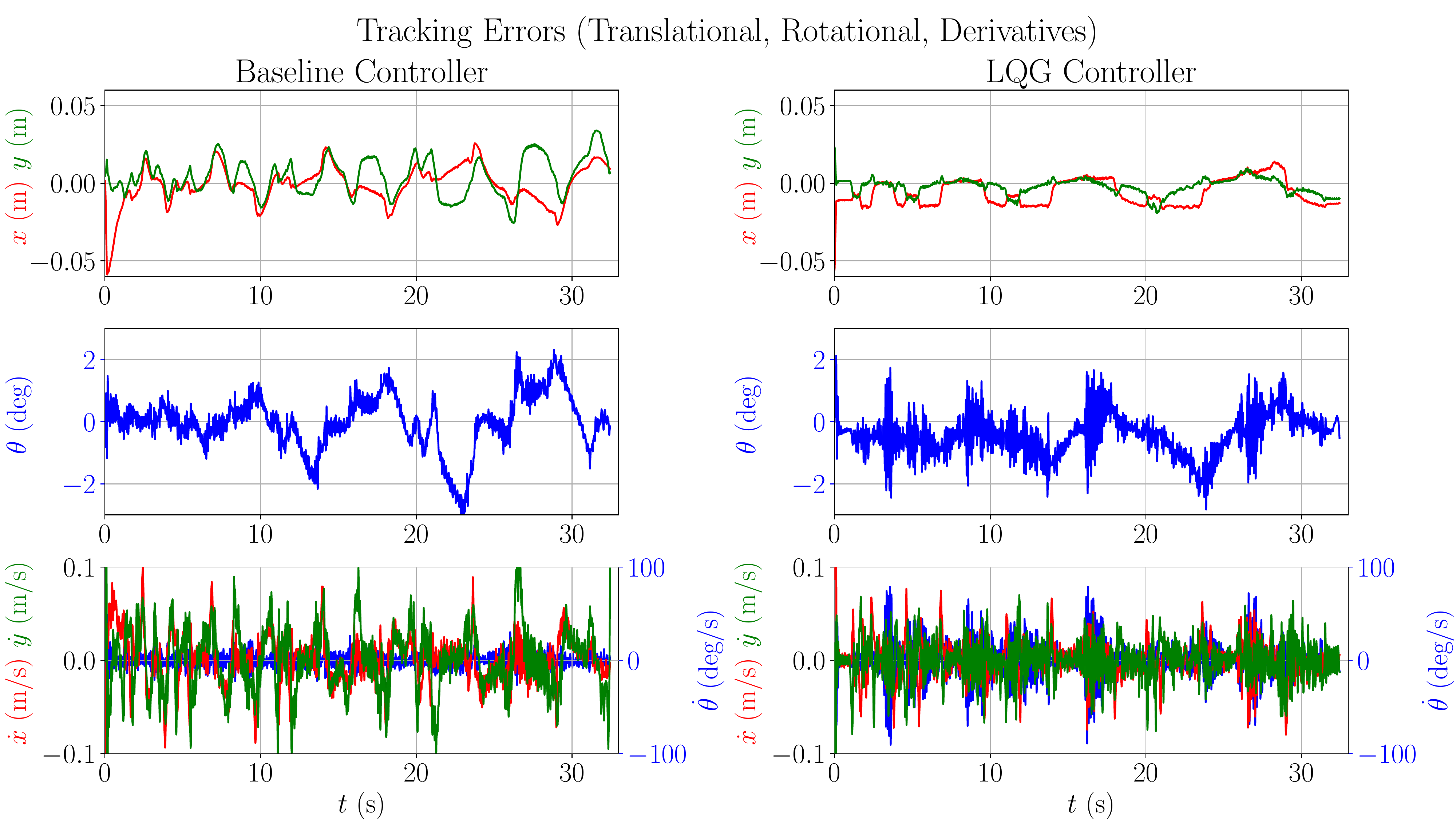}
  \caption{Tracking accuracy comparison between baseline (bottom left) and our proposed controller (bottom right) for a test trajectory (top).}
  \label{fig:experiment_1}
\end{figure}

Fig. \ref{fig:experiment_1} and Table \ref{tab:experiment_1} compare the results of our controller
against the baseline controller.  Note that, due to the robot geometry defined by the application in
\cite{Chen22icra_GTGraffiti}, orientation has limited control authority.  The baseline and LQG
controllers exhibit similar magnitudes of vibration around 9.9Hz and 13.1Hz, respectively, but
neither is qualitatively perceptible.

\begin{table}
  \centering
  \caption{Tracking Accuracies Near Workspace Center\\(root mean square deviation from $\xd$)}
  \begin{tabular}[t]{c|S|S}
    Dimension & {Baseline} & {LQG} \\ \hline \rule{0pt}{1.0\normalbaselineskip}
    $\theta$ (\SI{}{\deg}) & 0.9 & 0.8 \\
    $x$ (\SI{}{\mm}) & 12.4 & 10.3 \\
    $y$ (\SI{}{\mm}) & 12.8 & 5.4 \\
    $\dot{\theta}$ (\SI{}{\deg\per\s}) & 15.2 & 26.1 \\
    $\dot{x}$ (\SI{}{\mm\per\s}) & 44.3 & 36.1 \\
    $\dot{y}$ (\SI{}{\mm\per\s}) & 44.3 & 33.4
  \end{tabular}
  \label{tab:experiment_1}
\end{table}





%% file: 4-discussion.tex
The results in Section \ref{sec:experiments} demonstrate that our TV-LQG controller can perform favorably compared to the baseline state of the art PID-based controller.


Furthermore, we claim that, subjectively, the parameter tuning experience is much easier for our TV-LQG controller since most parameters (noise parameters) are measured and the LQR gain matrices $Q$ and $R$ are more intuitive and forgiving that tuning gain matrices directly -- indeed, LQR is often referred to as a technique for tuning gains \cite{Korayem12ijars_smc_lqr,Korayem17jirs_FeedbackLinearization_LQG}.
It can be shown that TV-LQG varies the gain matrices to maintain fixed pole/zero locations (for each linearized system, and for fixed $Q,R$) across the state space, though a full discussion is omitted for space reasons.

Because our approach relies heavily on offline computation, convergence and stability can be performed prior to execution.
Before any trajectory is run, the trajectory and controller can be checked numerically for feasibility and stability using optimality conditions and standard Lyapunov functions for LTV systems, respectively.  Failure to satisfy feasibility or stability conditions is an early indication that the desired trajectory or Q/R matrices must be modified to be realizable.

Additionally, we believe that our proposed TV-LQG controller is easier to implement since the embedded controller code consists of only 3 matrix multiplications and 3 vector additions per update.
The rest of the computation is in a high-level programming language without time constraints.

Finally, the modularity afforded by the factor graph framework, which is highly general, allows designers to add a wide variety of additional constraints, sensors, and control objectives with minimal modification.  Adding general constraints is described in \cite{Yang21icra_ecLQR}, adding sensors is a common usage of GTSAM for sensor fusion, and adding additional control objectives such as signed distance field obstacles factors have been demonstrated in \cite{Mukadam19ar_steap,Mukadam18ijrr_gpmp2}.

%% file: 5-conclusions.tex
In this work, we demonstrated how factor graphs can be used as a powerful tool to compute the optimal trajectory given quadratic state and cost objectives, compute the time varying LQR gains, and compute the time varying Kalman Filter Gains for optimal control of a redundantly actuated CDPR.
We also showed how the time varying Kalman Filter and LQR are combined into an online TV-LQG controller.
Finally, we successfully demonstrated our controller on a 4-cable planar CDPR and showed that it performs better across various regions in the workspace as compared to a baseline state-of-the-art dual-space feedforward controller.
Qualitatively, we believe that the factor graph approach is easier to understand and implement, and that our optimal control formulation is easier to tune than the baseline controller.

Future works include spatial SE(3) CDPR control; parameter bias drift modeling; online parameter estimation and replanning using Bayes Trees for incremental solving (estimation using \emph{smoothing} instead of \emph{filtering}, and model predictive control respectively); additional sensor and control modes (e.g. tension feedback); and additional non-ideality modeling such as cable stretch and sag.

%% file: 6_appendix.tex
The geometric, inertial, and friction parameters of the robot are listed in Table \ref{tab:robot_parameters}.  The geometry of the cable mounting points on both the fixed frame and movable end effector were directly measured, while the winch radiuses were estimated by a calibration procedure (described later).  The translational inertia was measured directly with a scale and the rotational inertia estimated using a CAD model.  The friction parameters were estimated by measuring winch no-load acceleration-velocity curves.  Friction due to pulleys was assumed to be negligible compared to motor and winch frictions.  Both controllers use the $F_s\tanh(\mu \ldot)$ model for static friction and we choose $\mu=$\SI{0.19}{\second\per\radian} so that the static friction is modeled to reach $0.9F_s$ by roughly \SI{0.1}{\m\per\s} ($\tanh(\SI{0.19}{\second\per\radian}\cdot\SI{7.87}{\radian\per\second})\approx 0.9$).

Calibration of winch radiuses is done by regressing the inverse kinematic solution against the logged motor encoder values.  This is done by recording the end effector pose and motor angles for an example trajectory.  Then, inverse kinematics is used to compute the ground truth length of the cable as the distance between the cable mounting location on the winch and the pulley location.  Finally, a linear regression is used to estimate winch radius and fine-tune the zero-position of the motor.

To tune the parameters for our LQG controller, we measured or estimated the covariances $\covmeas, \covdyn, \cov_0$ and gave an intuitive guess for the LQR $Q$ and $R$ weighting matrices.
We measured $\covmeas$ and $\covdyn$ by comparing the cable lengths, cable velocities, and end effector accelerations/motor torques between ground-truth and calibrated sensor measurements, so that the covariances are representative of the ``noise'' expected during runtime.  We assigned $\cov_0$ to be 0.1m as the initial placement error.
We assigned $R = I_{6x6}$ and $Q = \diag([10^2 ~ 10^4 ~ 10^4 ~ 0 ~ 0 ~ 0])$ which, intuitively, scales 1Nm of torque to be comparable to \SI{0.57}{\degree} and \SI{0.1}{\mm} (\SI{e-2}{\radian}, \SI{e-4}{\m}) of rotation and translation error respectively while ignoring velocity tracking error.

To tune the gains $K_p, K_i, K_d$ for our baseline dual-space feedforward controller, we applied manual methods.  We increased $K_p$ to a reasonable value without oscillations or shaking, then increased $K_d$ until oscillations occurred, increased $K_i$ until oscillations occurred, and finally fine-tuned the three.  Although we attempted using well-known strategies such as Ziegler-Nichols or halving the ultimate gain, we found that those gains were unusable.